\documentclass[letterpaper, 10 pt, journal, twoside]{IEEEtran}

\IEEEoverridecommandlockouts                              

\usepackage{amsmath} 

\usepackage{amssymb}  
\usepackage{bbm}
\usepackage[bb=boondox]{mathalfa}
\usepackage{mathtools}
\usepackage{graphicx} 
\usepackage{booktabs,arydshln}
\usepackage{array}
\makeatletter
\def\adl@drawiv#1#2#3{%
	\hskip.5\tabcolsep
	\xleaders#3{#2.5\@tempdimb #1{1}#2.5\@tempdimb}%
	#2\z@ plus1fil minus1fil\relax
	\hskip.5\tabcolsep}
\newcommand{\cdashlinelr}[1]{%
	\noalign{\vskip\aboverulesep
		\global\let\@dashdrawstore\adl@draw
		\global\let\adl@draw\adl@drawiv}
	\cdashline{#1}
	\noalign{\global\let\adl@draw\@dashdrawstore
		\vskip\belowrulesep}}
\makeatother

\usepackage{threeparttable}
\usepackage{tabularx}
\usepackage{multirow}
\usepackage[caption=false, font=footnotesize]{subfig}
\usepackage{algorithmic}
\usepackage{algorithm}

\usepackage{color}
\usepackage{xcolor}
\usepackage{colortbl}
\definecolor{green}{rgb}{0.8,1.0,0.8}
\definecolor{red}{rgb}{1.0,0.8,0.8}

\usepackage{comment}
\PassOptionsToPackage{hyphens}{url}
\usepackage{hyperref}
\usepackage[%
	backend=biber,
	url=true,
	block=space,
	maxbibnames=10,
	minbibnames=1,
	bibstyle=ieee,
]{biblatex}
\makeatletter
\def\blx@maxline{77}
\makeatother

\AtBeginBibliography{\footnotesize}
\AtEveryBibitem{
	\clearfield{doi}
	\clearfield{isbn}
	\clearfield{issn}
	\clearfield{month}
	\clearfield{publisher}
	\ifentrytype{inproceedings}{
		\clearfield{arxivId}
		\clearfield{archivePrefix}
		\clearfield{eprint}
		\clearfield{url}
		\clearfield{volume}
	}{}
	\ifentrytype{article}{
		\clearfield{arxivId}
		\clearfield{archivePrefix}
		\clearfield{eprint}
		\clearfield{url}
	}{}
	\ifentrytype{report}{
	}{}
}
\renewbibmacro{in:}{%
	\ifentrytype{inproceedings}{%
		\setunit{}
		\addperiod\addspace In \textit{Proc.\ of the}}%
	{\printtext{\bibstring{in}\intitlepunct}}
}
\DeclareSourcemap{
	\maps{
		\map{ 
			\step[fieldsource=url,
				match=\regexp{\{\\\_\}|\{\_\}|\\\_},
				replace=\regexp{\_}]
		}
		\map{ 
			\step[fieldsource=url,
				match=\regexp{\{\$\\sim\$\}|\{\~\}|\$\\sim\$},
				replace=\regexp{\~}]
		}
		\map{ 
			\step[fieldsource=url,
				match=\regexp{\{\\\x{26}\}},
				replace=\regexp{\x{26}}]
		}
	}
}
\usepackage{xspace}
\expandafter\def\expandafter\normalsize\expandafter{%
    \normalsize%
    \setlength\abovedisplayskip{3pt}%
    \setlength\belowdisplayskip{3pt}%
    \setlength\abovedisplayshortskip{3pt}%
    \setlength\belowdisplayshortskip{3pt}%
}

\begin{document}

\title{CLIP-Clique: Graph-based Correspondence Matching Augmented by Vision Language Models for Object-based Global Localization}

\author{Shigemichi Matsuzaki$^{1}$, \IEEEmembership{Member,~IEEE,}
	Kazuhito Tanaka$^{1}$, and Kazuhiro Shintani$^{1}$%
	\thanks{Manuscript received: April 7, 2024; Revised: June 27, 2024; Accepted: August 13, 2024.}
	\thanks{This paper was recommended for publication by Editor Cesar C. Lerma upon evaluation of the Associate Editor and Reviewers’ comments.}
	\thanks{$^{1}$S. Matsuzaki, K. Tanaka, and K. Shintani are with Frontier Research Center, Toyota Motor Corporation (TMC), Toyota, Aichi, Japan.
			{\tt\small shigemichi\_matsuzaki@mail.toyota.co.jp}
	}%
	\thanks{Digital Object Identifier (DOI): see top of this page.}
}

\markboth{IEEE ROBOTICS AND AUTOMATION LETTERS,  PREPRINT VERSION. ACCEPTED August 2024}%
{Matsuzaki \MakeLowercase{\textit{et al.}}: CLIP-Clique: Graph-based Correspondence Matching Augmented by Vision Language Models}


\maketitle
\begin{abstract}
	This letter proposes a method of global localization on a map with semantic object landmarks. One of the most promising approaches for localization on object maps is to use semantic graph matching using landmark descriptors calculated from the distribution of surrounding objects. These descriptors are vulnerable to misclassification and partial observations. Moreover, many existing methods rely on inlier extraction using RANSAC, which is stochastic and sensitive to a high outlier rate. To address the former issue, we augment the correspondence matching using Vision Language Models (VLMs). Landmark discriminability is improved by VLM embeddings, which are independent of surrounding objects. In addition, inliers are estimated deterministically using a graph-theoretic approach. We also incorporate pose calculation using the weighted least squares considering correspondence similarity and observation completeness to improve the robustness. We confirmed improvements in matching and pose estimation accuracy through experiments on ScanNet and TUM datasets.

\end{abstract}


\section{Introduction}
\IEEEPARstart{G}{lobal}
localization is a task where
the sensor pose relative to a prior map
is estimated using only a sensor observation
or a sequence of observations,
i.e., without prior information about its pose.
It can be applied to relocalization
for recovering from localization failure,
and loop closing
in Simultaneous Localization and Mapping (SLAM) systems.

Among vision-based mapping and localization approaches,
object-based methods enjoys
better robustness against severe viewpoint change
compared to traditional feature-based
visual SLAM like ORB-SLAM \cite{Mur-Artal2017},
which suffer from tracking failure
under large viewpoint changes \cite{Zins2022}, and
heterogeneity of cameras used
in mapping and localization phases.

In global localization in object-based maps,
the most straightforward yet widely used approach
is to establish correspondence candidates
by listing all possible pairs of
a landmark and an observation with the same object category,
followed by iterative hypothesis verification
such as Random Sample Consensus (RANSAC) \cite{Zins2022a}.
Such an approach is, however, sensitve to
outliers due to the stochasticity of the algorithm.
In our previous work \cite{Matsuzaki2024},
we employed a Vision Language Model (VLM)
to improve the efficiency and accuracy of
sample-based iterative correspondence matching.
While we showed its effectiveness,
the method still struggles with stably finding
the correct solution.

Some methods pose the problem of correspondence matching
as semantic graph matching \cite{Gawel2018a,Guo2021a}.
Unlike the aforementioned approaches,
where each correspondence candidate is treated individually
and spatial information is not considered,
those methods use object descriptors based on
the distributions of categories of surrounding objects.
Such descriptors encode a local spatial structure of
the landmarks and thus improve the discriminability of objects.
While these methods are promising, however,
descriptor matching 
may fail under partial observation and detection errors
because those factors change the distribution of surrounding objects.
In addition, they still rely on RANSAC 
to remove outliers after similarity-based correspondence matching.

\begin{figure}[tb]
	\centering
	\includegraphics[width=\hsize]{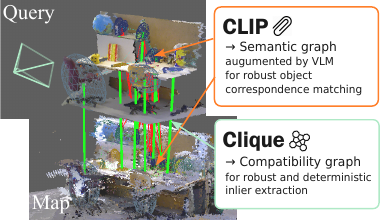}
	\vspace{-18pt}
	\caption{We propose \textbf{CLIP-Clique}, an object-based
		RGB-D global localization method
		driven by a novel correspondence matching strategy.
		It leverages two types of graphs: (i) 3D semantic graph
		for accurately estimating object correspondences, and
		(ii) spatial compatibility graph for efficiently
		extracting inlier correspondences as spatially compatible sets.
		We augment an existing semantic graph-based method \cite{Guo2021a}
		with a Vision Language Model, i.e., CLIP \cite{Radford2021}
		to enhance landmark discriminability and robustness.
		We also exploit CLIP-based similarity estimation
		in ranking multiple inlier candidates
		calculated as maximal cliques of the compatibility graph,
		and similarity-weighted least squares
		for accurate pose calculation.}
	\label{fig:top_image}
\end{figure}

In this letter, we introduce object correspondence matching
based on a combination of semantic graph and a VLM
to improve its robustness.
Here, we hypothesize that the landmark discriminability
in semantic graph matching
can be improved by neighbor-independent
object descriptors given by VLMs like CLIP \cite{Radford2021}.
We assign to each landmark an embedding vector from CLIP
and use it in similarity calculation between
observation and map objects.
In inlier correspondence extraction,
we employ a graph-theoretic method similar to \cite{Zhang2023c},
which builds a \textit{compatibility graph} that encodes
pairwise spatial consistency among the correspondence hypotheses,
and finds potential inlier sets as its maximal cliques.
To evaluate the likelihood of each candidate set,
we utilize the similarity of corresponding landmarks within the sets.
Lastly, we calculate a camera pose using the correspondences
via a weighted least squares method \cite{Malis2023}.
We design a weight of each correspondence considering
the correspondence similarity, and observation completeness,
i.e., coverage of observation over a single landmark.

To sum up, we improve each process of
semantic graph-based correspondence matching
and pose estimation using a VLM,
leading to our proposed method, coined \textit{CLIP-Clique}.
Our main contributions are as follows:
\begin{enumerate}
	\item Correspondence matching considering both
	      semantic histograms \cite{Guo2021a} and CLIP's semantic descriptors.
	\item A graph-theoretic extraction of multiple inlier sets
	      and similarity-based ranking to
	      determine more promising correspondences.
	\item Pose calculation via weighted least squares
	      based on correspondence similarity 
	      and observation completeness
	      to improve robustness to wrong correspondences
	      and incomplete observations.
\end{enumerate}


\section{Related Work}

\subsection{Object-based mapping and localization}

Object-based map representation has semantically meaningful
object instances as the central entities, i.e. landmarks \cite{Nicholson2019a}.
Leveraging the recent advancements in deep neural network (DNN)-based
object detectors like YOLO \cite{Wang2023i},
object-based mapping and localization provides
better robustness against viewpoint changes \cite{Yang2019b}.
Zins et al. \cite{Zins2022} introduced object-based relocalization to
supplement a feature-based visual SLAM \cite{Mur-Artal2017}
and improved its performance.

For global localization and relocalization in object-based maps,
many methods  rely on
RANSAC-like iterative approach on
matching candidates generated as all possible pairs
of observation and map landmarks with the same object category \cite{Zins2022,Zins2022a}.
Such a strategy of candidate generation inherently generates
many wrong correspondences (outliers).
It is known that RANSAC struggles with finding a true set of inlier correspondences
when the outlier ratio is high \cite{Yang2020g}. 
To mitigate this problem,
our previous work \cite{Matsuzaki2024} proposed
a global localization method named CLIP-Loc
using a VLM for more accurate correspondence matching,
and PROSAC \cite{Chum2005},  an improved iterative algorithm based on weighted sampling using
the correspondence similarity,
to improve the efficiency of the iterative inlier extraction.
While we saw significant improvements,
CLIP-Loc still relies on the unstable sampling-based algorithm.
In addition, the method only considers the similarity of
individual landmark-observation pairs and ignores
spatial distribution of the landmarks.

To improve the discriminability of object landmarks,
semantic graph-based methods
assign a descriptor to each object,
and match two graphs based on
the similarity of node descriptors.
In X-View \cite{Gawel2018a},
descriptors are calculated based on
patterns of object classes of nodes
visited via random walk.
Semantic Histogram (SH) proposed by
Guo et al. \cite{Guo2021a}
use histograms of object classes of nodes neighboring
the target node as descriptors.
While effective, 
those methods are vulnerable to
detection errors, partial observations,
and occlusions which affect
the connectivity of semantic graphs
and may deteriorate the descriptors.
Indeed, the SH \cite{Guo2021a} fails
when the observed semantic graph
differs from the map
as we demonstrate in Sec. \ref{sec:experiments}.
To complement this weakness,  
we incorporate a VLM in correspondence matching.

\subsection{Vision Language Models}

Vision Language Models (VLMs) \cite{Radford2021,Wang2022d}
are a type of large-scale machine learning models
capable of visual and textual tasks,
allowing for grounding visual information to
linguistic concepts.
VLMs has boosted the research of
various robotic tasks \cite{Huang2023c,Kawaharazuka2023a,Shridhar2021}.
CLIP \cite{Radford2021} is one of the most prominent
VLMs available at present.
It has separate encoders for text and image which
embed them in the common feature space
allowing multi-modal similarity estimation.

There are many studies that applied VLMs
in navigation tasks utilizing
text instructions and visual observations
\cite{Shah2022,Gadre2023}.
Several pieces work, e.g., \cite{Huang2023c},
embed CLIP features in the spatial representation
to enable text-based querying .
In the context of localization,
Mirjalili et al. proposed a method of scene recognition
based on image retrieval using
VLMs and LLMs \cite{Mirjalili2023}.
To the best of our knowledge,
CLIP-Loc \cite{Matsuzaki2024} is
the first method of object-based
global localization using VLMs.
In the present work, we extend \cite{Matsuzaki2024}
to RGB-D sensor-based global localization with
better object descriptors and the inlier extraction
strategy.

\subsection{Graph-based correspondence matching}

To extract correct correspondences from
a candidate set including outliers,
a traditional choice of algorithm is
Random Sample Consensus (RANSAC) \cite{Fischler1981} and its variants.
As the method is based on random sampling,
the result is inherently stochastic.
Moreover, the success rate of such methods quickly decreases
as the number of outliers 
increases \cite{Yang2020g}, because RANSAC assumes that
a set of correct correspondences are drawn by chance
out of the candidate set
within the pre-determined number of iterations.

As a different paradigm of correspondence matching,
graph-based methods have actively been studied in recent years
\cite{Yang2020g,Shi2021,Koide2022}.
This line of work establishes
\textit{compatibility graphs},
which encodes pair-wise consistency
between correspondence hypotheses,
and finds inliers as mutually connected sets
such as the maximum clique, maximal cliques,
etc.
Those methods are deterministic 
and thus more reliable.

The maximum clique-based algorithms \cite{Yang2020g,Shi2021,Koide2022}
assume there is only one true set of correspondences, 
and outliers are randomly distributed.
This is a strong assumption and does not hold when,
e.g., there are multiple likely hypotheses,
which can happen in localization problems.
Chen et al. \cite{Chen2022a} proposed 
using maximal clique finding.
Maximal cliques are complete subgraphs
that cannot be enlarged by including one more adjacent node,
and the maximum clique is the one with the most nodes.
\cite{Chen2022a} extracts multiple maximal cliques
and evaluates each resulting transformation.

We adopt the maximal clique-based
correspondence matching
to handle multiple correspondence hypotheses.
To estimate the likelihood of each hypothesis,
we employ similarity from both semantic histograms and
CLIP embeddings.

%
\begin{figure*}[tb]
	\centering
	\includegraphics[width=0.95\hsize]{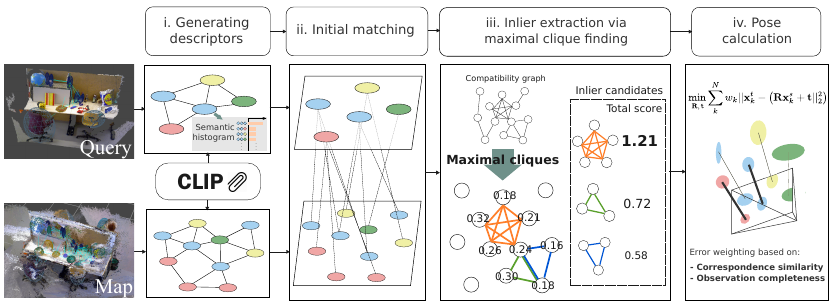}
	\vspace{-10pt}
	\caption{Overview of the proposed method.
		(i) For the given query and map landmarks,
		semantic graphs are built and the descriptors for each
		node is calculated. Specifically, we use
		Semantic Histograms \cite{Guo2021a} and CLIP embeddings
		as node descriptors (Sec. \ref{sec:proposed_method_object_descriptor}).
		(ii) Correspondence candidates
		are generated based on the similarity of
		corresponding object descriptors
		(Sec. \ref{sec:proposed_method_correspondence_generation}).
		(iii) From the initial correspondence set, inlier sets are
		extracted as sets of spatially compatible correspondences
		using the compatibility graph and maximal clique finding.
		Multiple sets are scored by the sum of the similarity
		to evaluate the likelihood 
		(Sec. \ref{sec:proposed_method_inlier_extraction}).
		(iv) A camera pose is calculated using the extracted inlier set.
		To mitigate the problem of wrong correspondences and incomplete observations,
		we employ weighted least squares based on
		the correspondence similarity and the observation completeness
		(Sec. \ref{sec:proposed_method_pose_estimation}).}
	\label{fig:overall_architecture}
\end{figure*}


\section{Problem definition}
\label{sec:proposed_method_overview}

Formally, we assume an object map $\mathcal{M}$
consisting of $N_m$ object landmarks, i.e.,
$\mathcal{M} = \{\mathbf{L}^{map}_m\}^{N_m}_{m=1}$,
$\mathbf{L}^{map}_m = \{\mathbf{Q}^{map}_m, l_m, c_m\}$,
where $\mathbf{Q}^{map}_{m}$, $l_m$, and $c_m$ denote
a dual form of a quadric fit to the landmark instance,
a text description, and an object class, respectively.
$\mathbf{Q}^{map}_m$ is
decomposed to the axis lengths $x_{map,m}, y_{map,m}, z_{map,m}$,
orientation $\mathbf{R}^{map}_{m}\in SO(3)$,
and position $\mathbf{t}^{map}_{m} \in \mathbb{R}^3$.
The text label $l_m$ describes the object's category,
attributes, appearance, etc. in a free-form text,
and given in arbitrary ways.

As an observation, we consider a single RGB-D frame consisting of
an RGB image $I$, and a depth image $D$.
We use an object detection model to get a set of $N_o$
detected objects $\{\mathbf{o}_n\}^{N_o}_{n=1}$,
each of which is represented as a tuple of
a bounding box, an object mask, and an object category label,
i.e., $\mathbf{o}_n=(\mathbf{b}_n, \mathbf{m}_n, c_n)$.
To reconstruct ellipsoidal object $\mathbf{Q}^{obs}_{n}$ for $\mathbf{o}_n$,
point cloud is generated by projecting depth values
within the mask $\mathbf{m}_n$, and a 3D oriented bounding box
is fitted via principal component analysis (PCA).
An ellipsoid is initialized with
the position, orientation, and axis lengths of
the 3D bounding box.
As a result, an observation object set
$\mathcal{O} = \{\mathbf{L}^{obs}_n\}^{N_o}_{n=1}$ is formed,
where $\mathbf{L}^{obs}_n = \{\mathbf{Q}^{obs}_n, \mathbf{b}_n, \mathbf{m}_n, c_n\}$.

To estimate the camera pose given a single query RGB-D observation,
we first establish correspondences between
$\mathcal{M}$ and $\mathcal{O}$,
and then calculate the 6-DoF pose.

%

\section{Proposed Method}
\label{sec:proposed_method}


The proposed method comprises the processes as follows:
1) generating object descriptors,
2) initial matching,
3) inlier extraction, and
4) pose calculation.
The overall pipeline is shown in Fig. \ref{fig:overall_architecture}.

Note that we employ two types of graph representation,
namely \textit{semantic graphs}
that describe the spatial and semantic information
of the objects, and
\textit{compatibility graphs}
that encode spatial consistency of object correspondence hypotheses,
described in Sec. \ref{sec:proposed_method_object_descriptor_sem_graph} and
\ref{sec:proposed_method_inlier_extraction_comp_graph}, respectively.

\subsection{Object descriptors}
\label{sec:proposed_method_object_descriptor}

As a descriptor for each object,
we combine CLIP \cite{Radford2021} and the Semantic Histograms \cite{Guo2021a}.

\subsubsection{CLIP descriptors}

For a map landmark $\mathbf{L}_m$,
a CLIP descriptor is calculated as a normalized embedding
of its text label $l_m$ from the text encoder:
\begin{equation}
	\mathbf{e}^{clip}_{map,m} = \text{CLIP}_{text}\left(l_m\right).
\end{equation}
Similarly, for an observed object $\mathbf{L}_n$,
a visual embedding is calculated as follows:
\begin{equation}
	\mathbf{e}^{clip}_{obs,n} = \text{CLIP}_{image}\left(\text{crop}\left(\mathbf{b}_n, I\right)\right),
\end{equation}
where $\text{crop}\left(\cdot, \cdot\right)$ denotes
an image cropping function.

\subsubsection{Semantic Histogram descriptors}
\label{sec:proposed_method_object_descriptor_sem_graph}

We adopt the Semantic Histograms (SH) \cite{Guo2021a}.
The object map and the observations are represented as
semantic graphs whose nodes represent
individual objects, and edges adjancency between two objects.
Two nodes are considered adjacent if
the distance between them is less than
a threshold $d_{adj}$ [m].

\begin{figure}[tb]
	\centering
	\includegraphics[width=0.55\hsize]{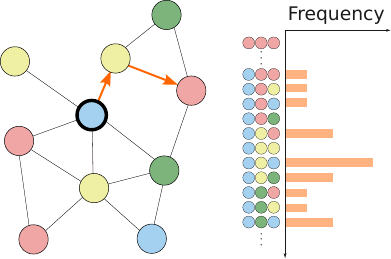}
	\vspace{-5pt}
	\caption{An illustration of the Semantic Histogram \cite{Guo2021a}.
		All possible paths with a fixed step length (here set to 3)
		starting from the target node are searched and the patterns of
		label sequences are recorded in a histogram.
		It effectively encodes the topological information around the object.}
	\label{fig:proposed_method_semantic_histogram}
\end{figure}

Next, a descriptor is generated for each node.
All possible paths with a fixed step length starting
from the target node are recorded
and the counts of label patterns are stored in a histogram
(see Fig. \ref{fig:proposed_method_semantic_histogram}).
The histogram is then L2-normalized to form
a $C^{s}$ dimensional descriptor,
where $C$ denotes the number of the classes,
and $s$ the step length (here $s=3$).

We refer to the resulting histogram descriptors for the map and observation objects
as $\mathbf{e}^{sh}_{map}$ and $\mathbf{e}^{sh}_{obs}$, respectively.

\subsubsection{Similarity score}

The similarity of a correspondence
between a map landmark and an observed landmark
is calculated as a weighted sum of the dot product of
the CLIP embeddings and that of SHs
assigned to the map and
the observation landmarks, i.e.,:
\begin{align}
	s_{total} & = \alpha s_{clip} +  (1-\alpha)s_{sh}  \label{eq:s_total},                 \\
	s_{clip}  & = \mathbf{e}^{clip}_{map} \cdot \mathbf{e}^{clip}_{obs}, \label{eq:s_clip} \\
	s_{sh}    & = \mathbf{e}^{sh}_{map} \cdot \mathbf{e}^{sh}_{obs}, \label{eq:s_sh}
\end{align}
where $\alpha \in [0, 1]$ is a weight coefficient. The effect of different values
of $\alpha$ is evaluated in Sec. \ref{sec:experiment_parameter_analysis}.

The similarity score in eq. \eqref{eq:s_total} encodes the class of objects and
their neighbors, as well as fine-grained appearance information.
Therefore, in the next initial matching,
the class is not used and only the similarity scores are considered.

\subsection{Initial matching}
\label{sec:proposed_method_correspondence_generation}

%
In the next step,
we generate a set of initial correspondence candidates,
which are later filtered in inlier extraction.
We first briefly review existing approaches
and their problems.

\textbf{Existing approaches}~
In \cite{Guo2021a}, the correspondence candidates are
generated as mutually optimal matches,
i.e, for each observation,
a landmark is considered a correspondence
if it has the highest similarity with the observation among all landmarks,
and vice versa.
CLIP-Loc \cite{Matsuzaki2024} instead uses
landmarks with $k$ nearest
descriptors 
as correspondence candidates 
to handle matching ambiguity.
We hereafter refer to those strategies as
\textit{1-to-1} and \textit{$k$NN} matching, respectively.

\textit{1-to-1} matching is 
excessively strict
and may lead to very sparse 
correspondences when
the observed semantic graph is corrupted.
Nonetheless, it is a reasonable choice to
keep the outlier rate as low as possible
for outlier-sensitve RANSAC.
\textit{kNN} loosens the 1-to-1 restriction to allow
multiple correspondence candidates considering
errors of the descriptors.
Although CLIP-Loc \cite{Matsuzaki2024}
employs PROSAC \cite{Chum2005}
to focus more on likely hypotheses,
the fixed number of candidates $k$
per observation cannot be too large
(set to $3$ in \cite{Matsuzaki2024}),
and it hinders incorporating candidates more than $k$.

\begin{figure}[tb]
	\centering
	\includegraphics[width=1.00\hsize]{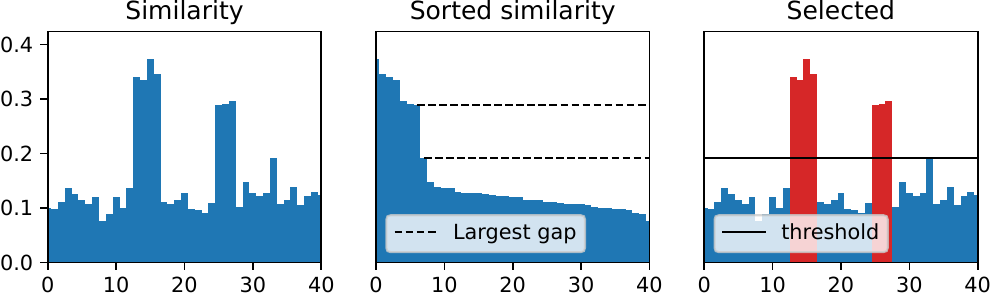}
	\vspace{-18pt}
	\caption{How to select correspondences for an observation.
		The horizontal axis of each subfigure corresponds the landmarks.
		When the similarity values are sorted,
		we set a threshold at the point where there is the largest gap
		to extract the arbitrary number of likely correspondence candidates.}
	\label{fig:matching_algorithm}
\end{figure}

\textbf{Our approach}~
To flexibly incorporate potential correspondences,
in this work, we use \textit{adaptive} matching strategy.
Intuitively, promising correspondences have
substantially larger similarity than others.
We identify such correspondences by finding
the largest similarity gap.
Fig. \ref{fig:matching_algorithm} visualizes the process.
The similarity values between the target observation and
all the landmarks are first calculated and sorted.
We then find the pair of consecutive values with
the largest difference.
The smaller value of the pair is used as a threshold $s_{thr}$.
In practice, we apply this algorithm
to top $M$ 
(e.g., a quarter of the landmarks)
of the sorted similarity values
to limit the maximum number of candidates. 
This method allows for picking
the different number of correspondences
based on the distribution of
similarity among the landmarks.




\subsection{Graph-theoretic inlier extraction}
\label{sec:proposed_method_inlier_extraction}

To robustly and deterministically
extract inlier correspondences from the outlier-contaminated set
found in Sec. \ref{sec:proposed_method_correspondence_generation},
we employ a graph-theoretic approach used in, e.g., \cite{Zhang2023c}.

\subsubsection{Building a compatibility graph}
\label{sec:proposed_method_inlier_extraction_comp_graph}

The graph-theoretic matching
approaches use a compatibility graph $\mathbf{C}$,
which encodes the local consistency
between every possible pair of correspondences,
expressed as an $N_{cand} \times N_{cand}$ matrix,
where $N_{cand}$ is the number of correspondence candidates.
Each graph node represents a correspondence,
and an edge is added
between two nodes that fulfill certain compatibility criteria.
Here, the compatibility is based on the rigid distance constraint \cite{Zhang2023c}.
For a pair $(\mathbf{c}_i, \mathbf{c}_j)$,
where $\mathbf{c}_i=(m, n)$ and $\mathbf{c}_j=(m', n')$ contain
indices of the map and observation landmarks,
the $ij$ element of $\mathbf{C}$ is set as follows:
\begin{align}
	\mathbf{C}_{ij} & = \begin{dcases}
		1 & d_{ij} < d_{comp}, i \neq j \\
		0 & \text{otherwise},
	\end{dcases}
\end{align}
where $d_{ij} = |d(\mathbf{t}^{map}_{m}, \mathbf{t}^{map}_{m'}) - d(\mathbf{t}^{obs}_{n}, \mathbf{t}^{obs}_{n'})|$,
$d(\mathbf{x}, \mathbf{y}) = ||\mathbf{x} - \mathbf{y}||_2$,
and $d_{comp}$ is the threshold of compatibility.


%
%
\subsubsection{Extracting compatible sets via maximal clique finding}

Next, we find sets of mutually compatible correspondences using
the compatibility graph.
The initial correspondence candidates generated in the previous step
include a true inlier set as well as wrong correspondences.
The wrong ones potentially include both complete outliers and
groups of consistent correspondences
such as structurally and semantically similar landmarks.
Maximum clique finding cannot flexibly handle such multi-modal solutions.
Therefore, here we employ inlier extraction strategy
using maximal cliques, inspired by \cite{Zhang2023c}.

We apply a maximal clique finding algorithm \cite{Bron1973}
to list multiple correspondence set hypotheses.
The hypotheses are sorted in descending order
by the sum of similarities of
correspondences within the set.
Top $N$ likely solutions can simply be yielded
as $N$ of the top of the sorted results.

\subsection{Pose estimation}
\label{sec:proposed_method_pose_estimation}

%
Given the extracted correspondence set
$\mathcal{C}=\{\mathbf{c}_k\}^{N_{c}}_{k=1}$,
calculation of the camera rotation $\mathbf{R}\in SO(3)$
and translation $\mathbf{t}\in\mathbb{R}^3$
is formulated as the following least-squares problem:
\begin{equation}
	\min_{\mathbf{R}, \mathbf{t}} \sum_{k}^{N_c} w_k ||\mathbf{t}^{map}_k - \left(\mathbf{R}\mathbf{t}^{obs}_k + \mathbf{t}\right)||^2_2,
	\label{eq:error_func}
\end{equation}
where, by abuse of notation,
$\mathbf{t}^{map}_{k}$ and $\mathbf{t}^{obs}_{k}$
are the positions of the ellipsoidal objects
of the map and observation in the correspondence $\mathbf{c}_k$, respectively.
A weight $w_k$ for each correspondence is calculated
based on the similarity and completeness of
observation, and defined as follows:
\begin{equation}
	w = w_{sim}w_{com}.
\end{equation}
$w_{sim}$ is a weight based on the similarity of
the corresponding nodes, i.e., $w_{sim} = s_{total}$.
$w_{com}$ is a completeness of the observation calculated as follows:
\begin{equation}
	w_{com} = min\left(1,\sqrt{\frac{x_{obs}^2 + y_{obs}^2 + z_{obs}^2}{x_{map}^2 + y_{map}^2 + z_{map}^2}}\right) \label{eq:w_com},
\end{equation}
where $x_{obs}, y_{obs}, z_{obs}$ are the length of the three axes of the observed ellipsoid,
and $x_{map}, y_{map}, z_{map}$ are those of the corresponding map ellipsoid, respectively.
When an object is observed completely,
the size of the observed ellipsoid is close to the map ellipsoid,
leading to $w_{com} \approx 1$.

We employ the closed-form solution for
a weighted least-squares problem
by Malis et al. \cite{Malis2023} to solve eq. \eqref{eq:error_func}.

%
\section{Experiments}
\label{sec:experiments}

\subsection{Setup}
\label{sec:experiments_setup}

\subsubsection{Implementation details}
\label{sec:experiments_setup_impl_details}

The proposed system and the baseline algorithms are implemented in Python 3.
We used the ViT-L/14 model of CLIP \cite{Radford2021},
open-sourced by OpenAI \protect\footnotemark.
\footnotetext{\url{https://github.com/openai/CLIP} (accessed on 26/3/2024)}
For maximal clique finding,
we used an algorithm by Bron and Kerbosch \cite{Bron1973}
implemented in NetworkX library \cite{Hagberg2008}.
The hyperparameters $d_{adj}$, $d_{comp}$, and $\alpha$
are empirically set to $0.8$ [m], $0.3$ [m], and $0.7$, respectively.
We conduct all experiments on a desktop computer with
a GeForce RTX 4090 GPU and a Intel Core i9 CPU (32 cores).

\subsubsection{Datasets}
\label{sec:experiments_setup_datasets}

We use 
ScanNet dataset \cite{Dai2017} and TUM RGB-D dataset \cite{Sturm2012}.
We choose ScanNet \textit{0002\_00} and \textit{0017\_00} as test sequences.
From TUM, we use \textit{fr2/desk} and \textit{fr3/long\_office\_household}.
In ScanNet, we use the ground-truth labels of instances and semantics
to generate observed landmarks.
In TUM, 
we used YOLOv8-x\protect\footnotemark \xspace
trained with COCO dataset \cite{Lin2014} to provide
instance-level observations.
\footnotetext{\url{https://github.com/ultralytics/ultralytics} (accessed on 26/3/2024)}

\textbf{Object maps} An object map with text labels is built for each sequence.
For ScanNet sequences, ellipsoidal landmarks are automatically reconstructed
and labeled with the object category IDs from the given labeled point cloud.
For TUM sequences, we use the manually built object maps used in \cite{Matsuzaki2024}.
The landmarks in all the maps are labeled with arbitrary description about the appearance
in English,
such as \textit{a yellow toy duck},  \textit{a purple office chair}, etc.

\subsubsection{Baselines}

\begin{table}[tb]
	\centering
	\caption{Algorithms used in the baselines and our method.}
	\label{table:experiment_baseline_params}
	\begin{threeparttable}
		{\scriptsize
			\begin{tabular}{ c c c c c }
				\toprule
				              & X-view \cite{Gawel2018a} & SH \cite{Guo2021a} & CLIP-Loc \cite{Matsuzaki2024} & \textbf{Proposed} \\
				\midrule
				Descriptor    & RW                       & SH                 & CLIP                          & SH+CLIP           \\
				Matching type & 1-to-1                   & 1-to-1             & $k$NN                         & adaptive          \\
				Inlier ext.   & RANSAC                   & RANSAC             & PROSAC                        & MC                \\
				\bottomrule
			\end{tabular}

			\begin{tablenotes}[para,flushleft,online,normal] 
				RW=random walk, SH=Semantic Histogram, MC=Maximal Clique finding\\
			\end{tablenotes}
		}
	\end{threeparttable}
\end{table}

We compare the proposed method with
two semantic graph-based baselines,
i.e., X-View \cite{Gawel2018a}, and
SH \cite{Guo2021a}.
We re-implemented the algorithms in Python 3.
In addition, we also use CLIP-Loc \cite{Matsuzaki2024}.
Note that those baselines use different descriptor types, matching strategies,
and inlier extraction methods, summarized in Table \ref{table:experiment_baseline_params}.
In RANSAC and PROSAC, the poses are verified
based on the total overlaps of the observations
and projection of landmarks used in \cite{Matsuzaki2024}.


%

\subsection{Comparison with the baselines}

\subsubsection{Correspondence matching} First,
we evaluate the performance of correspondence matching
using ScanNet dataset with the ground-truth instance labels.
Precision and recall are used as metrics. 
We use the estimated correspondence set
with the highest score calculated in Section \ref{sec:proposed_method_inlier_extraction}.

%
\begin{table*}[tb]
	\centering
	\caption{Precision and recall of correspondence matching, and pose estimation results on ScanNet and TUM RGB-D}
	\label{table:experiment_matching_accuracy}
	\begin{threeparttable}
		\begin{tabular}{ c c c c c c c c c c }
			\toprule
			\multirow{2}{*}{Sequence}      & \multirow{2}{*}{Method}        & \multirow{2}{*}{Precision [\%]$\uparrow$} & \multirow{2}{*}{Recall [\%]$\uparrow$} & \multirow{2}{*}{Trans. error [m] $\downarrow$ } & \multirow{2}{*}{Rot. error [rad] $\downarrow$ } & \multicolumn{3}{c}{Success rate [\%]$\uparrow$} & \multirow{2}{*}{Time [s]$\downarrow$}                                  \\
			\cmidrule(lr){7-9}
			                               &                                &                                           &                                        &                                                 &                                                 & Top 1                                           & Top 3                                 & Top 5         &                \\
			\midrule
			\multirow{4}{*}{\textit{0002}} & X-view \cite{Gawel2018a}       & 67.1                                      & 30.2                                   & 1.877                                           & 1.413                                           & 36.7                                            & 37.5                                  & 38.7          & 1.815          \\
			                               & SH \cite{Guo2021a}             & 70.1                                      & 33.7                                   & 1.709                                           & 1.294                                           & 39.8                                            & 40.4                                  & 41.3          & 1.743          \\
			                               & CLIP-Loc \cite{Matsuzaki2024}* & 37.5                                      & 11.8                                   & 2.379                                           & 1.820                                           & 19.6                                            & 30.0                                  & 36.5          & 1.888          \\
			                               & \textbf{Proposed}              & \textbf{75.9}                             & \textbf{36.9}                          & \textbf{0.772}                                  & \textbf{0.806}                                  & \textbf{72.9}                                   & \textbf{81.5}                         & \textbf{84.1} & \textbf{0.043} \\
			\midrule
			\multirow{4}{*}{\textit{0017}} & X-view                         & 65.1                                      & 17.6                                   & 2.244                                           & 1.197                                           & 24.8                                            & 24.8                                  & 25.6          & 1.090          \\
			                               & SH                             & \textbf{66.9}                             & 18.0                                   & 2.514                                           & 1.317                                           & 22.5                                            & 22.5                                  & 22.5          & 1.011          \\
			                               & CLIP-Loc*                      & 41.9                                      & 11.6                                   & 2.314                                           & 1.544                                           & 24.8                                            & 38.8                                  & 48.8          & 1.125          \\
			                               & \textbf{Proposed}              & 62.9                                      & \textbf{28.8}                          & \textbf{1.270}                                  & \textbf{1.125}                                  & \textbf{54.2}                                   & \textbf{67.1}                         & \textbf{71.4} & \textbf{0.041} \\
			\midrule
			\multirow{4}{*}{\textit{fr2}}  & X-view                         & -                                         & -                                      & 1.019                                           & 0.800                                           & 50.7                                            & 50.7                                  & 50.7          & 0.405          \\
			                               & SH                             & -                                         & -                                      & 0.888                                           & 0.661                                           & 62.6                                            & 62.6                                  & 62.6          & 0.400          \\
			                               & CLIP-Loc*                      & -                                         & -                                      & 0.701                                           & 0.485                                           & 83.2                                            & 92.3                                  & 94.8          & 0.631          \\
			                               & \textbf{Proposed}              & -                                         & -                                      & \textbf{0.529}                                  & \textbf{0.320}                                  & \textbf{91.1}                                   & \textbf{95.4}                         & \textbf{96.5} & \textbf{0.013} \\
			\midrule
			\multirow{4}{*}{\textit{fr3}}  & X-view                         & -                                         & -                                      & 1.249                                           & 0.945                                           & 44.9                                            & 44.9                                  & 44.9          & 0.714          \\
			                               & SH                             & -                                         & -                                      & 1.340                                           & 1.013                                           & 43.7                                            & 43.7                                  & 43.7          & 0.682          \\
			                               & CLIP-Loc*                      & -                                         & -                                      & 0.882                                           & 0.650                                           & 71.9                                            & 81.0                                  & 84.5          & 0.921          \\
			                               & \textbf{Proposed}              & -                                         & -                                      & \textbf{0.638}                                  & \textbf{0.482}                                  & \textbf{81.0}                                   & \textbf{86.3}                         & \textbf{88.9} & \textbf{0.042} \\
			\bottomrule
		\end{tabular}
		\begin{tablenotes}[para,flushleft,online,normal] 
			\item[*] CLIP-Loc uses an RGB observation.
		\end{tablenotes}
	\end{threeparttable}
\end{table*}

The results are shown in Table \ref{table:experiment_matching_accuracy}.
In ScanNet 0002, the proposed method significantly outperformed
the baselines in both precision and recall.
In 0017, although precision was worse than X-view and SH,
recall was better than them.
X-view and SH aim to increase the precision
by the strict 1-to-1 matching sacrificing the recall.
In contrast, the proposed method contributed to both metrics,
thanks to the combination of the good descriptors,
adaptive correspondence generation, and
the powerful graph-theoretic inlier extraction.

\subsubsection{Pose estimation} Next,
we evaluate the pose estimation performance.
We use success rate (SR), translation and rotation errors
(TE and RE, respectively) as metrics.
Success rates are calculated
as the ratio of the samples where
a pose is calculated with the translation error
less than 1.0 [m].

The proposed method outperformed the baselines by a large margin
in both sequences of ScanNet.
Although the average precision was not the best in 0017,
the final pose calculation results were significantly better.
Despite slightly low precision,
relatively higher recall suggests that
more correspondences were found.
It consequently increases the robustness to outliers.
We can expect this especially when the pose is calculated
with the proposed weighted least squares.

On TUM RGB-D, the our method also outperformed the baselines.
Despite the detections with errors by YOLOv8,
the method can accurately estimate the pose.
Top 1 to 5 results of X-view and SH
were the same because their 1-to-1 matching
gives only a single inlier set in most cases.
On the other hand,
ours handles multiple correspondences by
adaptive matching
and robustly finds consistent sets via the graph-theoretic method
resulting in gradually higher accuracy.

\subsubsection{Runtime} The proposed method required
no more than 0.05 seconds. 
Despite roughly $O(n^2)$ complexity of
the compatibility graph-based inlier extraction,
it is efficient as
is based on light-weight pair-wise compatibility evaluation.

\begin{figure}[tb]
	\centering
	\includegraphics[width=0.75\hsize]{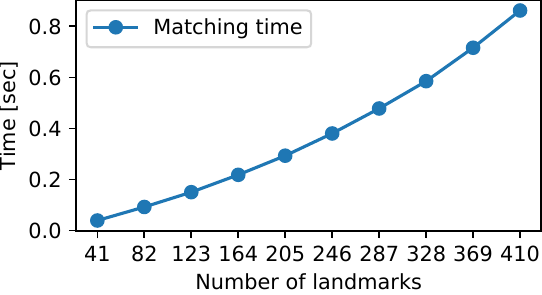}
	\vspace{-8pt}
	\caption{Relationship between the number of landmarks and the latency}
	\label{fig:experiment_scalability}
\end{figure}

To evaluate the scalability of the method,
we conducted additional experiments on
simulated scenes with more landmarks
by duplicating the original landmarks of \textit{fr3}.
The result is shown in Fig. \ref{fig:experiment_scalability}.
Although the growth of the computational time is quadratic,
the algorithm can handle about 410 landmarks within a second.
We can further optimize it via implementation in a faster language
and parallelization.

\subsection{Ablation studies}

To investigate the effect of each component in
the different steps of the pipeline,
we further evaluate the effect of
our proposals in each process.
We use the proposed CLIP-Clique as the base algorithm, 
and ablate the components.

\subsubsection{Object descriptors}

We compare different types of
object descriptors, i.e.,
\textit{SH} \cite{Guo2021a},
\textit{CLIP} \cite{Matsuzaki2024},
and the proposed hybrid method
of SH and CLIP (\textit{SH+CLIP}).
The results are shown in Table \ref{table:ablation_hypothesis_generation}.
The performance of \textit{SH} depended on the sequences, especially low in ScanNet 0017\_00.
Interestingly, \textit{CLIP} only also did not performed the best.
This may be because of CLIP's low accuracy on small observations, reported in \cite{Matsuzaki2024}.
In contrast, the hybrid descriptor (\textit{SH+CLIP}) consistently performed the best,
exploiting the complementary nature of both descriptors.

\begin{table}[tb]
	\centering
	\caption{Ablation study on object descriptors}
	\label{table:ablation_hypothesis_generation}
	{\scriptsize
		\begin{threeparttable}
			\begin{tabular}{ c c c c c c c }
				\toprule
				Seq.                           & Method                    & Pre. [\%]     & Rec. [\%]     & TE [m]         & RE [rad]       & SR [\%]       \\
				\midrule
				\multirow{3}{*}{\textit{0002}} & \textit{SH}               & 70.2          & 33.6          & 0.889          & 0.921          & 65.3          \\
				                               & \textit{CLIP}             & 37.9          & 11.8          & 1.433          & 1.424          & 38.5          \\
				                               & \textbf{\textit{SH+CLIP}} & \textbf{75.9} & \textbf{36.9} & \textbf{0.772} & \textbf{0.806} & \textbf{72.9} \\
				\midrule
				\multirow{3}{*}{\textit{0017}} & \textit{SH}               & 56.9          & 24.5          & 1.297          & 1.176          & 44.9          \\
				                               & \textit{CLIP}             & 41.4          & 16.7          & 1.745          & 1.529          & 37.2          \\
				                               & \textbf{\textit{SH+CLIP}} & \textbf{62.9} & \textbf{28.8} & \textbf{1.270} & \textbf{1.125} & \textbf{54.2} \\
				\midrule
				\multirow{3}{*}{\textit{fr2}}  & \textit{SH}               & -             & -             & 0.909          & 0.654          & 68.8          \\
				                               & \textit{CLIP}             & -             & -             & 0.628          & 0.410          & 84.7          \\
				                               & \textbf{\textit{SH+CLIP}} & -             & -             & \textbf{0.529} & \textbf{0.320} & \textbf{91.1} \\
				\midrule
				\multirow{3}{*}{\textit{fr3}}  & \textit{SH}               & -             & -             & 0.831          & 0.662          & 73.2          \\
				                               & \textit{CLIP}             & -             & -             & 0.870          & 0.660          & 74.5          \\
				                               & \textbf{\textit{SH+CLIP}} & -             & -             & \textbf{0.638} & \textbf{0.482} & \textbf{81.0} \\

				\bottomrule
			\end{tabular}
		\end{threeparttable}
	}
\end{table}

\begin{figure*}[tb]
	\centering
	\includegraphics[width=0.88\hsize]{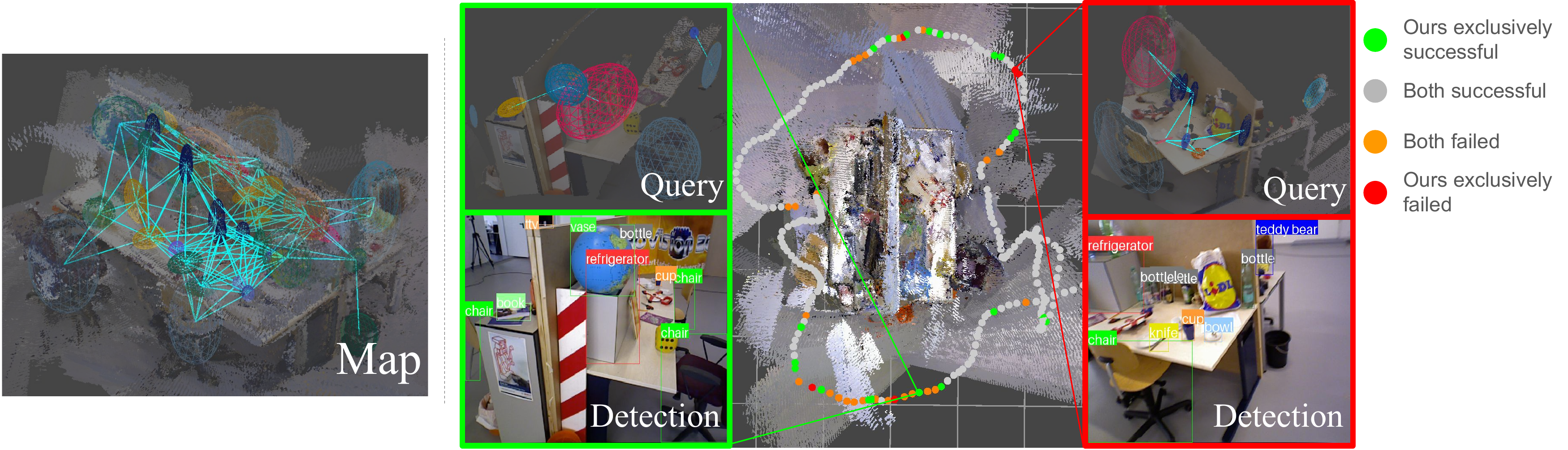}
	\vspace{-8pt}
	\caption{Visualization of the estimation results on
		TUM \textit{fr3/long\_office\_household}.
		\textbf{Left}: the object map and the semantic graph.
		\textbf{Right}: Top-down view of the sample trajectory,
		estimation results, and an exclusively successful and failure results
		compared to the SH \cite{Guo2021a}.
		In the successful case, the graph is way more sparse than the map,
		which will affect matching based on SHs.
		Nevertheless, pose estimation was successful thanks to CLIP-based descriptors.
		In the failure case, although there is more connectivity,
		correspondence matching failed.
		Looking at the detection, many observations are small and
		blurred. This might have affected the inference accuracy of CLIP,
		and led to failure.}
	\label{fig:experiment_exclusively_good_case}
\end{figure*}

To take a closer look at the advantage of the proposed method,
we examine the cases where only
the proposed method succeeded and failed to estimate accurate poses,
shown in Fig. \ref{fig:experiment_exclusively_good_case}.
In the exclusively successful case shown in green,
the query semantic graph was very sparse compared to the map
due to some objects missed by the detectors etc.
This deteriorates the performance of
semantic graph-only methods.
In contrast,
the proposed method successfully estimated the correspondences
and provided accurate results.
The method, however, also failed in cases
where the observations are small and with low quality,
presumably because, again, the limited accuracy of CLIP on small objects.

\subsubsection{Inlier extraction}

The results are shown in Table \ref{table:ablation_inlier_extraction}.
As baseline approaches to inlier extraction,
we use RANSAC, PROSAC, and maximal clique finding (\textit{Clique}).
For the stochastic iterative methods (RANSAC and PROSAC),
we report the results of the best success rate among
three trials.
The maximal clique-based inlier extraction consistently outperformed
RANSAC and PROSAC.
Moreover, the graph-based method is deterministic and
thus more reliable.

\begin{table}[tb]
	\centering
	\caption{Ablation study on inlier extraction methods}
	\label{table:ablation_inlier_extraction}
	{\scriptsize
		\begin{threeparttable}
			\begin{tabular}{ c c c c c c c }
				\toprule
				Seq.                           & Method                   & Pre. [\%]     & Rec. [\%]     & TE [m]         & RE [rad]       & SR [\%]       \\
				\midrule
				\multirow{3}{*}{\textit{0002}} & \textit{RANSAC}          & 67.6          & 18.2          & 1.920          & 1.454          & 37.3          \\
				                               & \textit{PROSAC}          & 74.9          & 19.6          & 1.865          & 1.411          & 39.6          \\
				                               & \textbf{\textit{Clique}} & \textbf{75.9} & \textbf{36.9} & \textbf{0.772} & \textbf{0.806} & \textbf{72.9} \\
				\midrule
				\multirow{3}{*}{\textit{0017}} & \textit{RANSAC}          & 55.3          & 15.4          & 1.809          & 1.146          & 41.1          \\
				                               & \textit{PROSAC}          & 58.5          & 16.2          & 1.772          & \textbf{1.102} & 44.0          \\
				                               & \textbf{\textit{Clique}} & \textbf{62.9} & \textbf{28.8} & \textbf{1.270} & 1.125          & \textbf{54.2} \\
				\midrule
				\multirow{3}{*}{\textit{fr2}}  & \textit{RANSAC}          & -             & -             & 0.573          & 0.381          & 88.9          \\
				                               & \textit{PROSAC}          & -             & -             & 0.578          & 0.389          & 89.3          \\
				                               & \textbf{\textit{Clique}} & -             & -             & \textbf{0.529} & \textbf{0.320} & \textbf{91.1} \\
				\midrule
				\multirow{3}{*}{\textit{fr3}}  & \textit{RANSAC}          & -             & -             & 0.758          & 0.570          & 76.1          \\
				                               & \textit{PROSAC}          & -             & -             & 0.776          & 0.591          & 75.4          \\
				                               & \textbf{\textit{Clique}} & -             & -             & \textbf{0.638} & \textbf{0.482} & \textbf{81.0} \\
				\bottomrule
			\end{tabular}

		\end{threeparttable}
	}
\end{table}

\subsubsection{Pose estimation}

We ablate the weight factors in the pose estimation
to evaluate their effect.
The results are shown in Table \ref{table:ablation_pose_estimation}.
\textit{none} means no weighting among the correspondences.
\textit{sim} and \textit{com} denote
weighting strategies using the correspondence similarity (eq. \eqref{eq:s_total})
and the completeness (eq. \eqref{eq:w_com}), respectively,
and \textit{both} uses both of them.
Although the effect is not significant, \textit{both} consistently resulted in
the best performance in most of the metrics.
As a component, \textit{com} contributed more to the accuracy of pose estimation.

\begin{table}[tb]
	\centering
	\caption{Ablation on weighting strategies in pose calculation}
	\label{table:ablation_pose_estimation}
	{\scriptsize
		\begin{threeparttable}
			\begin{tabular}{ c c c c c }
				\toprule
				Seq.                           & Method                 & TE [m] $\downarrow$ & RE [rad] $\downarrow$ & SR [\%]$\uparrow$ \\
				\midrule
				\multirow{4}{*}{\textit{0002}} & \textit{none}          & 0.802               & 0.837                 & 72.1              \\
				                               & \textit{sim}           & 0.799               & 0.832                 & 72.2              \\
				                               & \textit{com}           & 0.778               & 0.814                 & 72.6              \\
				                               & \textbf{\textit{both}} & \textbf{0.772}      & \textbf{0.806}        & \textbf{72.9}     \\
				\midrule
				\multirow{4}{*}{\textit{0017}} & \textit{none}          & 1.287               & 1.131                 & 54.5              \\
				                               & \textit{sim}           & 1.283               & \textbf{1.124}        & \textbf{55.0}     \\
				                               & \textit{com}           & 1.274               & 1.131                 & 54.0              \\
				                               & \textbf{\textit{both}} & \textbf{1.270}      & 1.125                 & 54.2              \\
				\midrule
				\multirow{4}{*}{\textit{fr2}}  & \textit{none}          & 0.545               & 0.334                 & 90.6              \\
				                               & \textit{sim}           & 0.542               & 0.331                 & 90.9              \\
				                               & \textit{com}           & 0.531               & 0.322                 & 90.9              \\
				                               & \textbf{\textit{both}} & \textbf{0.529}      & \textbf{0.320}        & \textbf{91.1}     \\
				\midrule
				\multirow{4}{*}{\textit{fr3}}  & \textit{none}          & 0.656               & 0.485                 & 79.7              \\
				                               & \textit{sim}           & 0.649               & \textbf{0.481}        & 80.2              \\
				                               & \textit{com}           & 0.644               & 0.485                 & 80.9              \\
				                               & \textbf{\textit{both}} & \textbf{0.638}      & 0.482                 & \textbf{81.0}     \\

				\bottomrule
			\end{tabular}

		\end{threeparttable}
	}
\end{table}

\subsubsection{Initial matching strategies}

We evaluate three strategies for initial correspondence generation,
\textit{1-to-1} used in \cite{Gawel2018a,Guo2021a},
i.e., \textit{$k$NN} used in \cite{Matsuzaki2024},
and \textit{adaptive} matching used in the proposed method.
The \textit{1-to-1} matching resulted in the worst performance
in matching accuracy, especially recalls,
due to its extremely strict nature.
\textit{$k$NN} provided better results than \textit{1-to-1}
and even marked the highest success rate in ScanNet 0017.
The \textit{adaptive} matching resulted in the best performance
especially in TUM sequences,
presumably because
TUM sequences have a different number of similar objects
such as 
bottles and chairs, which
can be better handled by the flexible correspondence generation
with the graph-based robust inlier extraction.

\begin{table}[tb]
	\centering
	\caption{Ablation study on initial matching strategies}
	\label{table:ablation_initial_matching}
	{\scriptsize
		\begin{threeparttable}
			\begin{tabular}{ c c c c c c c }
				\toprule
				Seq.                           & Method                     & Pre. [\%]     & Rec. [\%]     & TE [m]         & RE [rad]       & SR [\%]       \\
				\midrule
				\multirow{3}{*}{\textit{0002}} & \textit{1-to-1}            & 68.4          & 25.0          & \textbf{0.666} & \textbf{0.661} & 61.5          \\
				                               & \textit{kNN}               & 73.4          & 35.7          & 0.813          & 0.868          & 71.3          \\
				                               & \textbf{\textit{adaptive}} & \textbf{75.9} & \textbf{36.9} & 0.772          & 0.806          & \textbf{72.9} \\
				\midrule
				\multirow{3}{*}{\textit{0017}} & \textit{1-to-1}            & 60.6          & 18.7          & 1.529          & 1.323          & 35.2          \\
				                               & \textit{kNN}               & 61.3          & 28.3          & 1.272          & \textbf{1.109} & \textbf{55.2} \\
				                               & \textbf{\textit{adaptive}} & \textbf{62.9} & \textbf{28.8} & \textbf{1.270} & 1.125          & 54.2          \\
				\midrule
				\multirow{3}{*}{\textit{fr2}}  & \textit{1-to-1}            & -             & -             & 0.577          & 0.364          & 86.9          \\
				                               & \textit{kNN}               & -             & -             & 0.560          & 0.350          & 89.5          \\
				                               & \textbf{\textit{adaptive}} & -             & -             & \textbf{0.529} & \textbf{0.320} & \textbf{91.1} \\
				\midrule
				\multirow{3}{*}{\textit{fr3}}  & \textit{1-to-1}            & -             & -             & 0.727          & 0.560          & 76.7          \\
				                               & \textit{kNN}               & -             & -             & 0.709          & 0.535          & 78.3          \\
				                               & \textbf{\textit{adaptive}} & -             & -             & \textbf{0.638} & \textbf{0.482} & \textbf{81.0} \\
				\bottomrule
			\end{tabular}

		\end{threeparttable}
	}
\end{table}

\subsection{Parameter analysis}
\label{sec:experiment_parameter_analysis}

\begin{figure}[tb]
	\centering
	\includegraphics[width=0.85\hsize]{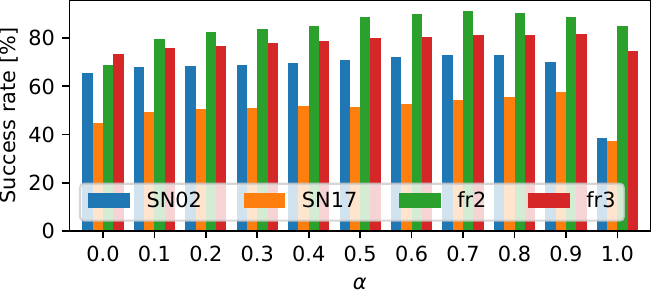}
	\vspace{-10pt}
	\caption{Parameter analysis for similarity weight $\alpha$ in eq. \eqref{eq:s_total}}
	\label{fig:evaluation_alpha}
\end{figure}

Lastly, we evaluate the effect of the weight value $\alpha$
on the CLIP-based similarity (see eq. \eqref{eq:s_total}).
Fig. \ref{fig:evaluation_alpha} shows the Top 1 success rates for
the four benchmark sets with different values of $\alpha$.
The best performance was achieved with $\alpha$ between $0.7$ and $0.9$ in all sequences.

\textbf{Why should $\alpha$ be large?}~
It is presumably due to the difference of actual value ranges that
the two similarity measures take.
In the experiments, the maximum value of CLIP similarity (eq. \eqref{eq:s_clip}) was
no more than 0.4, while SH similarity (eq. \eqref{eq:s_sh}) about 0.8.
This suggests that the optimal value of $\alpha$ is to roughly
equalize the two similarity values.

\textbf{Why does adding CLIP NOT affect the performance in ScanNet where CLIP performs poorly?}~
We empirically found that when CLIP results in poor accuracy,
the estimated similarity values among the landmarks
tend to be similar and the difference among them is negligible compared to
the value of SH similarity.
In such cases, eq. \eqref{eq:s_total} can be interpreted as adding
a constant to the SH similarity values.
We thus conclude that the hybrid use of CLIP and SH is always recommended
regardless of the performance of CLIP.
\section{Conclusions and Future Work}

In this letter, we proposed an object-based
global localization method coined CLIP-Clique.
The core of the proposal is
combining semantic graph-based matching \cite{Guo2021a}
and CLIP \cite{Radford2021}
to improve the matching accuracy and robustness,
coupled with the graph-theoretic inlier extraction
for better stability and accuracy.
The final pose estimation accuracy was also
improved by weighted least squares
considering the correspondence similarity
and observation completeness.

As a next step,
we are looking to apply this method
in relocalization of visual SLAM systems
for robust recovery
similar to \cite{Zins2022}, and object-based
loop closing.




%

%



\printbibliography

\end{document}